\documentclass{article}
\usepackage{ijcai22}

\usepackage{times}
\usepackage{soul}
\usepackage{url}
\usepackage[hidelinks]{hyperref}
\usepackage[utf8]{inputenc}
\usepackage[small]{caption}
\usepackage{graphicx}
\usepackage{amsmath}
\usepackage{amsthm}
\usepackage{amssymb}
\usepackage{booktabs}
\usepackage{algorithm}
\usepackage{algorithmic}
\usepackage{multirow}
\title{Noise-Robust Bidirectional Learning with Dynamic Sample Reweighting\footnote{Runner-up of the 1st Learning and Mining with Noisy Labels Challenge in IJCAI-ECAI 2022. This is an informal technical report.}}

\author{
Chen-Chen Zong
\and
Zheng-Tao Cao
\and
Hong-Tao Guo
\and
Yun Du
\and 
Ming-Kun Xie
\and
\\ Shao-Yuan Li
\and
Sheng-Jun Huang
\affiliations
College of Computer Science and Technology, Nanjing University of Aeronautics and Astronautics \\
MIIT Key Laboratory of Pattern Analysis and Machine Intelligence, Nanjing, 211106
\emails
\{chencz,caozhengtao,csguo, muyun, mkxie, lisy,huangsj\}@nuaa.edu.cn
}

\begin{document}

\maketitle

\begin{abstract}
    Deep neural networks trained with standard cross-entropy loss are more prone to memorize noisy labels, which degrades their performance. Negative learning using complementary labels is more robust when noisy labels intervene but with an extremely slow model convergence speed. In this paper, we first introduce a bidirectional learning scheme, where positive learning ensures convergence speed while negative learning robustly copes with label noise. Further, a dynamic sample reweighting strategy is proposed to globally weaken the effect of noise-labeled samples by exploiting the excellent discriminatory ability of negative learning on the sample probability distribution. In addition, we combine self-distillation to further improve the model performance. The code is available at \url{https://github.com/chenchenzong/BLDR}.
\end{abstract}

\section{Introduction}

Supervised learning largely depends on large and carefully curated labeled datasets. However, accurately labeling a large number of examples is daunting and time-consuming, inevitably introducing some noisy labels. This label noise degrades the performance of deep neural networks (DNNs) when directly trained with standard cross-entropy loss, and it consequently raises a significant challenge to learn with noisy labeled data.

Existing approaches for robust learning with noisy labels can be roughly divided into the following four categories: 1) Learning with Noise Transition \cite{xiao2015learning,chen2015webly,bekker2016training,sukhbaatar2014training}. 2) Sample Reweighting \cite{liu2015classification,chang2017active,zhang2021dualgraph}. 3) Self/Co-Training \cite{han2018co,yu2019does,jiang2018mentornet}. 4) Robust Loss Functions \cite{ghosh2017robust,zhang2018generalized,wang2019symmetric,lyu2019curriculum,feng2021can}. Thereinto, the treatment of noisy labeled samples mainly includes: reducing their contribution in the loss, correcting their label, or abstaining their classification.

Complementary labels are labels other than the given labels. Recently, negative learning using complementary labels has been shown more robust to cope with noisy labels. Specifically, samples with noisy label will receive a larger penalty and thus can be better distinguished from clean labeled samples on the sample probability distribution. However, the lack of information contained in the complementary labels makes convergence extremely slow when training the model using negative learning, hindering its further blooming.

Unlike previous negative learning-based methods that directly identify noisy labeled samples by setting a threshold and correct their labels, we introduce the corrected labels into the negative learning loss and adopt a reweighting strategy for overall training by normalizing the predicted probability of the model trained by negative learning. Further, we modify the model and propose a bidirectional learning scheme in which adding a positive learning (eg. standard cross-entropy) head to ensure convergence speed. In addition, we use SOP+ \cite{liu2022robust} as the baseline method and combine self-distillation to further improve the model performance. The effectiveness of the proposed method is demonstrated by multiple experiments conducted on the CIFAR-10N and CIFAR-100N datasets for image classification and label noise detection. For simplicity, we use BLDR (\textbf{B}idirectional \textbf{L}earning with \textbf{D}ynamic Sample \textbf{R}eweighting) to refer to the proposed method.

Our contribution can be summarized into three aspects:

\begin{itemize}
	\item We propose a bidirectional learning scheme by combining positive and negative learning, which makes the model more robust to noise and still maintain a fast model convergence speed.
	
	\item To further weaken the effect of noise-labeled samples, we introduce a dynamic sample reweighting strategy by normalizing the predicted probability of the model trained by negative learning and propose a new way to use pseudo labels generated by two model heads.
	
	\item We extend the proposed method with SOP+ and a self-distillation method proposed in \cite{zhang2019your}. Multiple experiments conducted on the CIFAR-10N and CIFAR-100N datasets for image classification and label noise detection can demonstrate the effectiveness of our method.
	
\end{itemize}

\section{Proposed Method}

\begin{figure}
	\centering
	\includegraphics[width=0.49\textwidth]{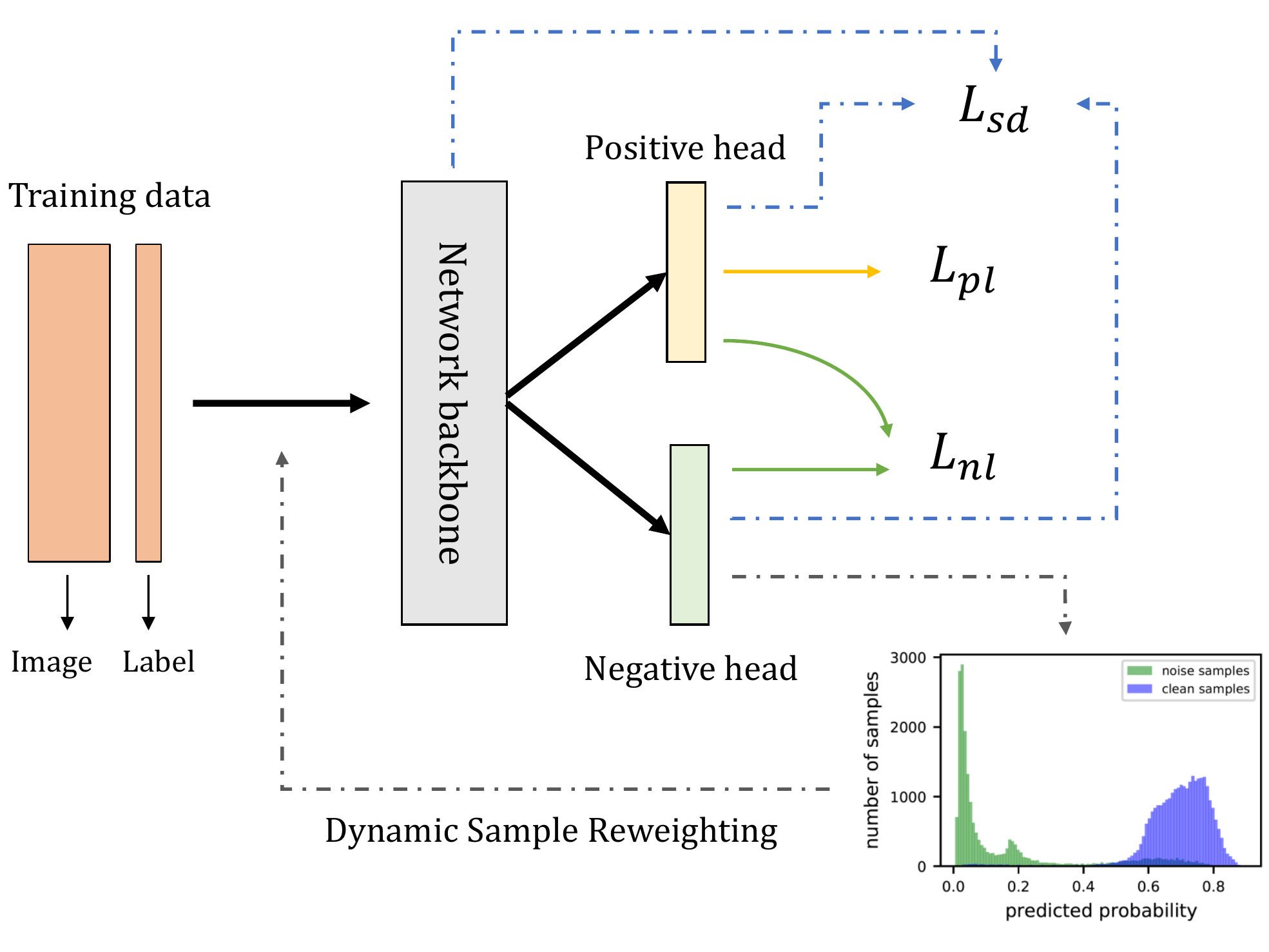}
	\caption{Overview of the BLDR method. We adopt a two-head model, which has a positive head trained in the normal way and a negative head trained with Equation \ref{nl}. By normalizing the predicted probability of the negative head, we assign a weight to each sample.}
	\label{fig.1}
\end{figure}

We consider the problem of ordinary multi-class classification. Let $x\in \mathcal{X}$ be an input, $y, \bar{y} \in \mathcal{Y} =\left \{ 1,...,c \right \} $ be its clean label and noisy label respectively, wherein $c$ is the size of labels. Suppose the CNN $f(x,\theta )$ maps the input space to the $c$-dimensional label space $f:\mathcal{X} \to \mathbb{R} ^c$, where $\theta$ denotes the network parameters. The goal of label-noise learning is to learn a classifier $f$ by minimizing the classification risk: $\mathcal{R} (f)=\mathbb{E}_{p(x,\bar{y})}l(f(x,\theta ), \bar{y} )$, where $p(x,\bar{y})$ denotes the noise distribution.

Figure \ref{fig.1} shows an overview of our method which we describe next.

\subsection{Bidirectional Learning Scheme}

In multi-class classification problems, the cross-entropy loss is the most commonly used loss function, which is expressed as:

\begin{equation} \label{ce}
\mathcal{L}_{ce}(f,\bar{y} ) =-\sum_{k=1}^{c}\bar{y} _k\log_{}{p_k},
\end{equation}

where $p$ is the predicted probability by $f$ and $p_k$ denotes the $k^{th}$ element of $p$. However, previous work has shown that training neural networks directly using standard cross entropy loss are more prone to memorize noisy labels and therefore degrade their performance. To cope with this problem, a large body of work has attempted to modify the loss function to make it more robust to noisy data. Among them, negative learning as an indirect learning scheme that uses complementary labels has recently gained increasing attention due to its special tolerance for noisy labeled data. Specifically, the negative learning loss function takes the following form:

\begin{equation} \label{nl}
\mathcal{L}_{nl}(f,\hat{y} ) =-\sum_{k=1}^{c}\hat{y}  _k\log_{}{(1-p_k)},
\end{equation}

where $\hat{y}$ is the complementary label which is a label other than the given label $y$. Obviously, noisy labeled samples will be assigned to the correct label with a specific probability and thus suffer a greater penalty. Eventually, samples with noisy label can be better distinguished from clean labeled samples on the sample probability distribution.

Unfortunately, complementary labels contain less information compared to the given labels, which leads to the fact that directly training using complementary labels will result in slow convergence or even non-convergence of the target model. To ensure convergence of the model, \cite{wang2021learning} proposes to combine Equation \ref{ce} and Equation \ref{nl} and regulate them with one extra parameter. This is indeed a feasible approach, but we find that doing so weakens the model's ability to discriminate between noisy and clean labeled samples.

In a similar way, we also try to combine Equation \ref{ce} and Equation \ref{nl} but in a completely different way. Inspired by representation learning, we adopt a two-head model structure, where one head (positive head) is trained using Equation \ref{ce} to ensure model convergence speed and the other head (negative head) is trained using Equation \ref{nl} to preserve robustness to noise. Since the two intersect at the representation layer, they can contribute to each other but do not affect the head's ability to discriminate against noise.

\subsection{Dynamic Sample Reweighting Strategy}


\begin{figure}
	\centering
	
	\includegraphics[width=4cm]{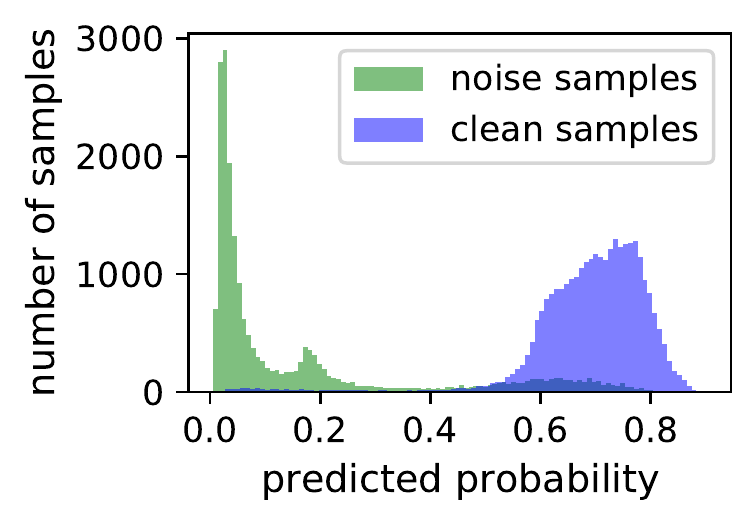}
	\includegraphics[width=4cm]{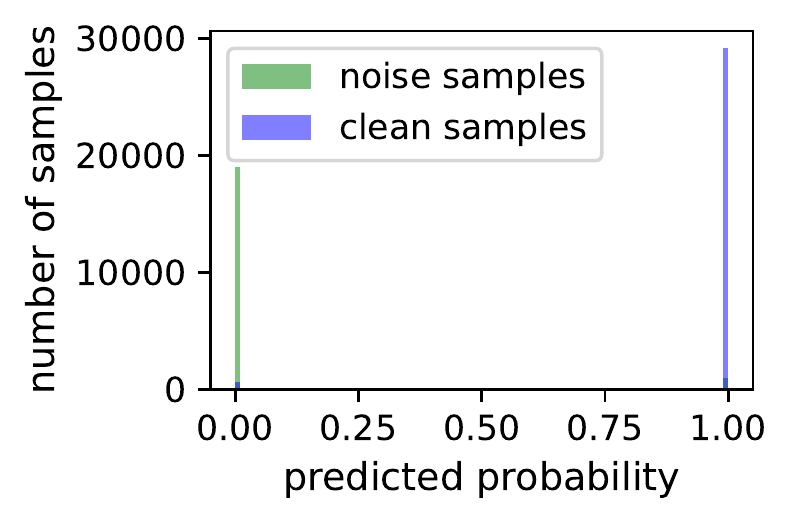}
	
	\caption{The predicted probability statistics for all CIFAR-10N-Worst training samples on the noisy labels. Figure left only uses bidirectional learning scheme to train the model, while right one combines with the dynamic sample reweighting strategy.}
	\label{fig.2}
\end{figure}

Figure \ref{fig.2} left shows the statistics of the predicted probabilities by negative head for all CIFAR-10N-Worst \cite{wei2021learning} training samples on the noisy labels. Most of the previous approaches classify samples into clean and noisy samples by setting a threshold, and then train the model in a semi-supervised way. However, we can see from the Figure \ref{fig.2} that there is still a portion of noisy samples entangled with clean samples in the high predicted probability region which inevitably hinder the performance of previous methods. 

We believe that noisy samples with high predicted probability have confusing characteristics or coexist with multiple target categories. Obviously, samples with confusing characteristics are hard to identify.
To handle samples coexist with multiple target categories, we adjust the definition of $\hat{y}$ in negative learning: $\hat{y}$ is a label other than the given label $y$ and the corrected label $\tilde{y} $ for every iteration during training, wherein $\tilde{y}$ is the predicted class by adding the output of two heads.

Further, by normalizing the predicted probability of all samples, we can assign a weight to each sample. Specifically, the weights are generated dynamically during the training process, thereinto the current round weights are derived from the previous round prediction results and the weights of all samples are initially $1$. The possibly mislabeled samples tend to be assigned small weights, while large weights tend to assign the potentially clean samples. Eventually, the trained model will have excellent noise tolerance, as shown in Figure \ref{fig.2} right.


\subsection{Self-distillation Loss and SOP+}

To further improve the model performance, we introduce a self-distillation method proposed in \cite{zhang2019your}. The method adds multiple output heads in the middle layers of the model and its loss function consists of three main components:

\begin{itemize}
	\item The standard training loss (eg. cross-entropy) from labels to all the shallow heads. 
	
	\item The KL (Kullback-Leibler) divergence loss under teacher’s guidance. The deepest head is considered as the teacher and the shallow heads are students.
	
	\item L2 loss between features maps of the deepest head and each shallow head.
	
\end{itemize}

Suppose we introduce $t$ new output heads $\left \{ f_1, f_2,...,f_t\right \} $ in the middle layers. Use $\left \{ p^1, p^2,..., p^t\right \} $ and $p_{ens}$ denote the predicted probablility from $\left \{ f_1, f_2,...,f_t\right \} $ and $f$, respectively. $\left \{ F_1, F_2,...,F_t\right \} $ and $F_{pl}$ are used to denote the features in $\left \{ f_1, f_2,...,f_t\right \} $ and $f_{pl}$, respectively. The self-distillation losse can be expressed as:

\begin{equation} \label{sd}
	\begin{aligned}
\mathcal{L}_{sd} = & \sum_{j}^{t}  CrossEntropy(p^j, \bar{y} ) + \alpha \sum_{j}^{t}KL(p^j,p_{ens})
\\& +\lambda \sum_{j}^{t}\left \| F_j-F_{pl} \right \| _{2}^{2}.   
\end{aligned}
\end{equation}

Task-specific, in the competition we add SOP+ \cite{liu2022robust} to the positive head as a regularization term. SOP+ also consists of three parts: standard SOP, consistency regularization and class-balance regularization. Hence, $\mathcal{L}_{pl}$ can be expressed as:

\begin{equation} \label{pl2}
	\mathcal{L}_{pl}  = \mathcal{L}_{ce} + \beta \mathcal{L}_{SOP} + \gamma \mathcal{L}_{consistency} + \delta \mathcal{L}_{class}.  
\end{equation}

 The final overall loss is :

\begin{equation} \label{all}
\mathcal{L} = \mathcal{L}_{pl} + \mathcal{L}_{nl} + \mathcal{L}_{sd}. 
\end{equation}

\section{Experiments}

\subsection{Experiment Setup}

Our experiments are conducted on two datasets: CIFAR-10N with aggre, rand1 and worst noise cases; CIFAR-100N with noisy fine case. We choose image classification and label noise detection to test our proposed method separately.

\textbf{Implementation Details:} The network structure of our approach is based on PreAct ResNet34 and we modify the network according to \cite{zhang2019your}. Positive and negative heads have the same structure, both are a single layer fully connected network. In experiments on CIFAR-10N (CIFAR-100N), we train the model with 320 (340) epochs, where the first 20 (40) epochs are used as a warm-up without sample reweighting. SGD is adopted as the optimizer with momentum 0.9, weight decay 5e-4, initialization learning rate 0.04, and batch size of 256. We use CosineAnnealingLR as the learning rate scheduler with T-max 300 and eta-min 2e-4. For the weight parameters in the loss, \{$\alpha, \lambda , \gamma, \delta$\} are set to \{0.1, 1e-6 , 0.9, 0.1\}, while $\beta$ is defined as the $r^2 \times 50$, assuming $r$ denotes the estimated noise ratio. To estimate $r$ and complete label noise detection task, a threshold $h$ is needed to classify clean and noisy samples, for which $h$ we set to 0.3 in all experiments. We use exponential moving average (EMA) to update the model and save the model in the last epoch.

\subsection{Image Classification}

Table \ref{table:acc} shows the comparison of our method with several other methods that have been shown to be effective on the CIFAR-N dataset. It can be observed that the proposed BLDR can outperform other previous works by a large margin on CIFAR-10N with aggre, rand1 and worst noise cases. For time reasons, on CIFAR-100N we follow the parameter settings on CIFAR-10N, which may not be ideal.

Considering that both PES(semi) and Divide-Mix, which use a semi-supervised idea for noise learning, achieve good results on CIFAR-100N. This is due to the fact that CIFAR-100N has only 500 training samples per category which is much lower than CIFAR-10N's 5000, and a portion of useful information is lost when learning by reweighting samples or adding regularization term. Combining our approach with semi-supervision is what we hope to do subsequently.

\subsection{Label Noise Detection}

Table \ref{table:f1} shows the label noise detection precision, recall and F1 score of our method calculated by setting a threshold to split clean samples and noisy samples. It is worth noting that for all noise cases of CIFAR-10N and CIFAR-100N, the set thresholds are the same. 

\begin{figure}
	\centering
	\includegraphics[width=0.45\textwidth]{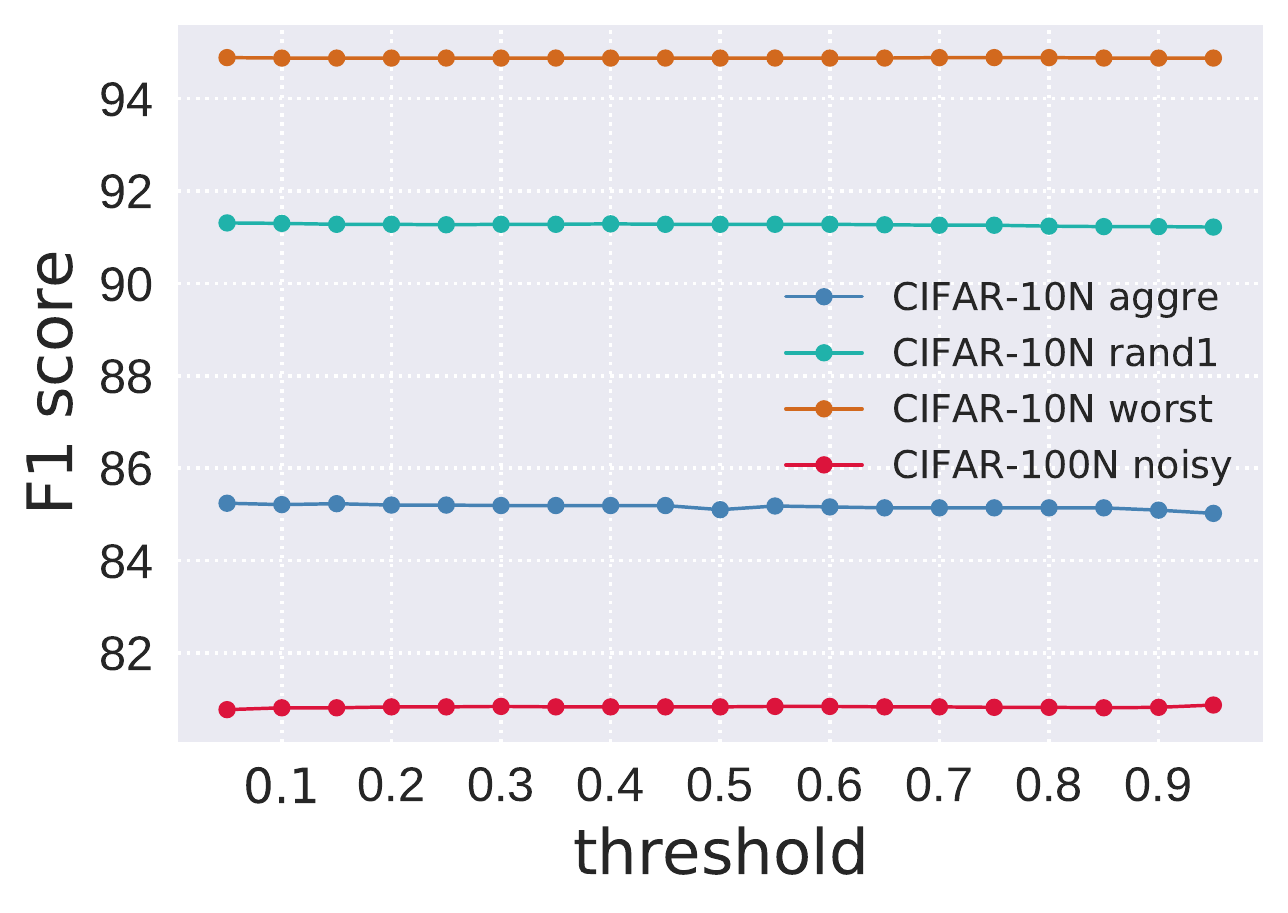}
	\caption{The relationship between the threshold value and the final F1 score obtained among different datasets and different noise cases.}
	\label{fig.3}
\end{figure}

Figure \ref{fig.3} shows the relationship between the threshold value and the final F1 score obtained. Referring to Figure \ref{fig.2} right, our method does not need to deliberately choose the threshold value even for different datasets or different noise cases due to its powerful noise discrimination capability.

\begin{table*}[!t]\small	
	\renewcommand\arraystretch{1.5}
	\centering
	\caption{Comparison of test accuracies (\%) on CIFAR-10N and CIFAR-100N using different methods. The results of the comparison methods are derived from \url{http://noisylabels.com/}, and we bold the best case of the comparison methods to visually compare with BLDR. $\uparrow $ indicates the boost value of our method, no boost is indicated by "-".}
	\label{table:acc}{
		\begin{tabular}{cc|c|c|c|c|c|c|c}
			\hline
			\multicolumn{2}{c|}{Method} & \multicolumn{1}{c}{CORES*} & \multicolumn{1}{|c}{PES(semi)}
			&\multicolumn{1}{|c|}{ELR+} &\multicolumn{1}{c}{Divide-Mix} 
			&\multicolumn{1}{|c}{SOP} &\multicolumn{1}{|c}{BLDR} &\multicolumn{1}{|c}{$\uparrow $}\\
			
			\hline
			
			\multirow{3}*{CIFAR-10N} & \multicolumn{1}{|c|}{Aggre} & $95.25\pm 0.09$ & $94.66\pm 0.18$ & $94.83\pm 0.10$ & $95.01\pm 0.71$ & $\pmb{95.61\pm 0.13}$ & $\pmb{96.45}$ & $0.84$ \\
			\cline{2-9}
			
			& \multicolumn{1}{|c|}{Rand1} & $94.45\pm 0.14$ & $95.06\pm 0.15$ & $94.43\pm 0.41$ & $95.16\pm 0.19$ & $\pmb{95.28\pm 0.13}$ & $\pmb{96.20}$ & $0.92$ \\
			\cline{2-9}
			
			& \multicolumn{1}{|c|}{Worst} & $91.66\pm 0.09$ & $92.68\pm 0.22$ & $91.09\pm 1.60$ & $92.56\pm 0.42$ & $\pmb{93.24\pm 0.21}$ & $\pmb{95.16}$ & $1.92$ \\
			
			\hline
			
			\multirow{1}*{CIFAR-100N} & \multicolumn{1}{|c|}{Noisy} & $55.72\pm 0.42$ & $70.36\pm 0.33$ & $66.72\pm 0.07$ & $\pmb{71.13\pm 0.48}$ & $67.81\pm 0.23$ & $-$ & $-$  \\
			
			\hline
	\end{tabular}}
\end{table*}

\begin{table*}[!t]\small	
	\renewcommand\arraystretch{1.5}
	\centering
	\caption{The label noise detection performance of our method. A fixed threshold is set for the various noise cases on CIFAR-N.}
	\label{table:f1}{
		\begin{tabular}{cc|ccc|c}
			\hline
			\multicolumn{2}{c|}{Dataset} & \multicolumn{3}{c}{CIFAR-10N}
			&\multicolumn{1}{|c}{CIFAR-100N} \\
			
			\hline
			\multicolumn{2}{c|}{Noisy type} & Aggre & Rand1 & Worst & Noisy \\
			\hline
			\multirow{3}*{BLDR} & \multicolumn{1}{|c|}{Precision} & $84.13$ & $89.50$ & $96.78$ & $-$  \\
			\cline{2-6}
			
			& \multicolumn{1}{|c|}{Recall} & $89.63$ & $94.29$ & $95.08$ & $-$  \\
			\cline{2-6}
			
			& \multicolumn{1}{|c|}{F1} & $86.79$ & $91.83$ & $95.92$ & $-$  \\
			
			\hline
	\end{tabular}}
\end{table*}

\section{Conclusions}

In this paper, we propose a noise-robust bidirectional learning scheme with dynamic sample reweighting, BLDR for short. BLDR adopts a two-head network structure to combine positive and negative learning in the representation layer. By doing so, the convergence speed of the model is guaranteed and the model is made to have excellent noise tolerance. Further, we try to weaken the interference of noisy data by normalizing the predicted probabilities of the training samples by negative head on the noisy markers as weights. In addition, we have combined self-distillation and SOP+ methods with our method and achieved satisfactory results in both image classification and label noise detection on CIFAR-N datasets.

\bibliographystyle{named}
\bibliography{ijcai22}

\end{document}